\documentclass[runningheads]{llncs}

\usepackage[T1]{fontenc}
\usepackage{graphicx}
\usepackage{tikz}
\usepackage{amsmath}
\usepackage{amssymb}
\usepackage{hyperref}
\usepackage{booktabs}
\usepackage{multirow}
\usepackage{bbding}
\usepackage{pifont}

\begin{document}

\title{Semantic Allocation in Ordered Bottlenecks: Predictive Residual Inference for Visual Representation Learning}
\titlerunning{Semantic Allocation in Ordered Bottlenecks}

\author{Erik Ayari\Envelope\orcidID{0009-0003-9408-9762} \and Manuel Traub\orcidID{0000-0003-0897-1701} \and Martin V. Butz\orcidID{0000-0002-8120-8537}}
\authorrunning{E. Ayari et al.}
\institute{Neuro-Cognitive Modeling Group, University of T{\"u}bingen, T{\"u}bingen, Germany\\
\email{erik.ayari@uni-tuebingen.de}\\
\email{manuel.traub@uni-tuebingen.de, martin.butz@uni-tuebingen.de}}

\maketitle

\begin{abstract}
	Ordered bottlenecks aim to provide utility at flexible budgets by assigning coarse information to early tokens and task-relevant detail to later ones.
	Prior work, including tail dropping (TD), typically enforces ordering by means of a masking-based ordering pressure (MBOP):
	Late tokens are masked more frequently than early tokens and are therefore encouraged to store less essential fine details.
	We introduce predictive residual inference for ordered representations (PRIOR), a framework designed to address inherent weaknesses of MBOP.
	MBOP is prone to weak late-token utility because it lacks an explicit refinement objective and uses gradient exposure as a proxy for importance.
	Furthermore, representations may become particularly brittle in optimization-sensitive settings, such as when using discrete or quantized token representations.
	PRIOR replaces activation-rate control with $\log_2$-scaled levels and level-wise predictors.
	These predictors separate already explained from unexplained information, focusing each level on residual error.
	We compare PRIOR against MBOP-TD and independent tail-biased dropout (MBOP-ITD) in contrastive learning and image reconstruction tasks.
	Unlike the baselines, PRIOR learns well-ordered representations across experiments: low budgets provide coarse descriptors, while high budgets add refinements.
	Simultaneously, full-budget performance with PRIOR is higher in all but one experimental setting, where performance remains comparable.
	MBOP baselines are severely limited in discrete and quantized settings, while PRIOR approaches the performance of continuous counterparts.
	Taken together, these findings establish PRIOR as an effective framework for ordered representation learning.
\keywords{predictive residual inference \and ordered representations \and tail dropping \and contrastive learning \and coarse-to-fine processing}
\end{abstract}

\section{Introduction}
Most deep learning architectures treat latent representations as fixed-size codes: During training and inference, the full code is produced, transmitted, and consumed by downstream components.
This assumption becomes restrictive when computational, storage, or communication resources vary.
With limited bandwidth, a mobile system may initially transmit only coarse scene information, and load additional details only after a user expresses interest.
Similarly, an embodied agent may first obtain rough object estimates, select task-relevant objects, and then request finer details only for those.
To support such adaptive use cases, representations should be hierarchically structured such that early tokens provide compact descriptors, while later tokens add \textit{complementary} information when richer inference is needed.

Common methods for learning ordered representations impose order through masking or truncation during training, thereby inducing a masking-based ordering pressure (MBOP).
For example, nested dropout stochastically removes contiguous sets of hidden units, making earlier dimensions more important for reconstruction and retrieval \cite{rippel2014learning}.
Related rateless auto-encoding approaches use gradually varying dropout rates to support flexible representation sizes \cite{koike2020stochastic}.
More recently, quality-controllable image tokenizers have used tail dropping (TD) to concentrate important information near the beginning of a discrete token sequence \cite{miwa2025one}.
Together, these methods show that masking and truncation can turn an otherwise flat latent sequence into a useful prefix code.

Despite their effectiveness, MBOP methods have notable weaknesses.
First, the ordering pressure is tied directly to gradient exposure: later positions are masked more often and therefore receive weaker training signals.
As a result, increasing the representation budget may yield diminishing returns, undermining the goal of budget-controllable representations.
Second, reduced gradient exposure may be especially problematic for representations with biased or high-variance gradient estimates, such as categorical one-hot or quantized vectors.
Finally, MBOP encourages ordering only implicitly, without an explicit objective that trains later tokens to refine the content already provided by earlier ones.

Models of human cognition offer a useful complementary perspective.
Perception is not only a bottom-up accumulation of local features.
Instead, global structure can influence perception from early on.
Accordingly, the visual system can be viewed as a hierarchical processing system, in which coarse prior knowledge about scenes or objects activates top-down expectations that facilitate perception \cite{bar2006topdown,hochstein2002view,navon1977forest}.
Predictive coding models make this idea computationally explicit: higher levels predict lower-level activity in a top-down manner, while feedforward signals transmit residual prediction errors from lower to higher levels \cite{rao1999predictive}.
These findings suggest a representational principle: priors predict coarse structural outlines, while additional capacity should encode residual information, that is, details the coarser representation cannot explain.

We introduce PRIOR, a framework for learning hierarchical representations with flexible operating points.
PRIOR separates ordering pressure from gradient exposure by learning a $\log_2$-scaled hierarchy of token sequences.
From coarse to fine, self-predictive modules double the latent resolution at each level and use residual errors to explicitly recruit the added capacity for information not yet explained by coarser levels.

We evaluate PRIOR across training objectives, downstream tasks, and architectural choices, comparing it against two MBOP baselines: standard tail dropping (TD) and independent tail-biased dropout (ITD).
The first experimental track uses instance-level temporal contrastive learning and evaluates class readouts at multiple hierarchy levels.
The second track studies reconstruction quality in an autoencoding task.

\section{Methods}
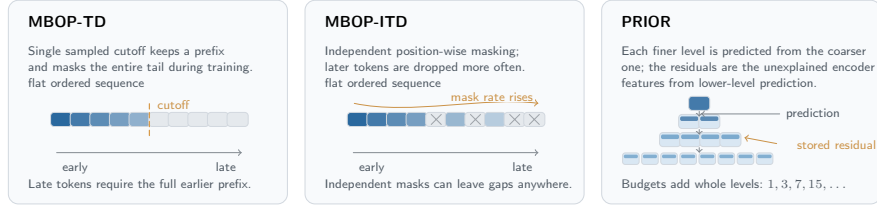
\begin{figure}[ht]
\centering
\resizebox{\linewidth}{!}{
%

\definecolor{oaBlue}{RGB}{42,104,158}
\definecolor{oaBlueSoft}{RGB}{109,161,201}
\definecolor{oaBluePale}{RGB}{217,231,243}
\definecolor{oaSlate}{RGB}{78,88,98}
\definecolor{oaBorder}{RGB}{210,216,223}
\definecolor{oaPanel}{RGB}{248,250,252}
\definecolor{oaMuted}{RGB}{229,233,237}
\definecolor{oaWarm}{RGB}{226,162,79}

\begin{tikzpicture}[x=1cm,y=1cm,line cap=round,line join=round]
  \tikzset{
    oa panel/.style={
      draw=oaBorder,
      fill=oaPanel,
      rounded corners=2mm,
      line width=0.45pt
    },
    oa title/.style={
      font=\sffamily\bfseries\small,
      text=black
    },
    oa subtitle/.style={
      font=\sffamily\scriptsize,
      text=oaSlate,
      align=left
    },
    oa note/.style={
      font=\sffamily\scriptsize,
      text=oaSlate,
      align=left
    },
    oa callout/.style={
      font=\sffamily\scriptsize,
      text=oaSlate,
      align=left,
      fill=oaPanel,
      inner xsep=1.2pt,
      inner ysep=0.6pt
    },
    oa chip/.style={
      draw=oaBorder,
      fill=white,
      rounded corners=1.2mm,
      inner xsep=1.7mm,
      inner ysep=0.8mm,
      font=\sffamily\scriptsize,
      text=oaSlate
    },
    oa token/.style={
      draw=oaBorder,
      rounded corners=0.7mm,
      line width=0.35pt,
      minimum width=0.42cm,
      minimum height=0.28cm,
      inner sep=0pt
    },
    oa tiny token/.style={
      draw=oaBorder,
      rounded corners=0.55mm,
      line width=0.30pt,
      minimum width=0.33cm,
      minimum height=0.23cm,
      inner sep=0pt
    },
    oa arrow/.style={
      ->,
      draw=oaSlate!70,
      line width=0.45pt
    },
    oa warm arrow/.style={
      ->,
      draw=oaWarm!90!black,
      line width=0.55pt
    }
  }

  \def\panelw{5.55}
  \def\panelh{4.05}
  \def\panelgap{0.35}

  \begin{scope}[shift={(0,0)}]
    \draw[oa panel] (0,0) rectangle (\panelw,\panelh);

    \node[oa title, anchor=west] at (0.28,3.63) {MBOP-TD};
    \node[oa subtitle, anchor=west] at (0.28,3.02)
      {Single sampled cutoff keeps a prefix};
    \node[oa subtitle, anchor=west] at (0.28,2.70)
      {and masks the entire tail during training.};
    \node[oa subtitle, anchor=west] at (0.28,2.38)
      {flat ordered sequence};

    \node[oa token, fill=oaBlue] at (1.05,1.68) {};
    \node[oa token, fill=oaBlue!88!white] at (1.44,1.68) {};
    \node[oa token, fill=oaBlue!76!white] at (1.83,1.68) {};
    \node[oa token, fill=oaBlue!64!white] at (2.22,1.68) {};
    \node[oa token, fill=oaBlue!52!white] at (2.61,1.68) {};
    \node[oa token, fill=oaMuted] at (3.00,1.68) {};
    \node[oa token, fill=oaMuted] at (3.39,1.68) {};
    \node[oa token, fill=oaMuted] at (3.78,1.68) {};
    \node[oa token, fill=oaMuted] at (4.17,1.68) {};
    \node[oa token, fill=oaMuted] at (4.56,1.68) {};

    \draw[densely dashed, draw=oaWarm!95!black, line width=0.70pt]
      (2.805,1.36) -- (2.805,2.02);
    \node[oa callout, anchor=west, text=oaWarm!90!black] at (2.92,1.98)
      {cutoff};

    \draw[oa arrow] (0.96,1.02) -- (4.64,1.02);
    \node[oa note, anchor=north west] at (0.96,0.94) {early};
    \node[oa note, anchor=north east] at (4.64,0.94) {late};

    \node[oa note, anchor=north west, text width=4.95cm] at (0.28,0.60)
      {Late tokens require the full earlier prefix.};
  \end{scope}

  \begin{scope}[shift={({\panelw+\panelgap},0)}]
    \draw[oa panel] (0,0) rectangle (\panelw,\panelh);

    \node[oa title, anchor=west] at (0.28,3.63) {MBOP-ITD};
    \node[oa subtitle, anchor=west] at (0.28,3.02)
      {Independent position-wise masking;};
    \node[oa subtitle, anchor=west] at (0.28,2.70)
      {later tokens are dropped more often.};
    \node[oa subtitle, anchor=west] at (0.28,2.38)
      {flat ordered sequence};

    \draw[oa warm arrow] (0.98,2.00) .. controls (2.10,1.76) and (3.60,2.14) .. (4.64,2.00);
    \node[oa callout, anchor=east, text=oaWarm!90!black] at (4.58,2.14)
      {mask rate rises};

    \node[oa token, fill=oaBlue] at (1.05,1.68) {};
    \node[oa token, fill=oaBlue!88!white] at (1.44,1.68) {};
    \node[oa token, fill=oaBlue!76!white] at (1.83,1.68) {};
    \node[oa token, fill=oaBlue!64!white] at (2.22,1.68) {};
    \node[oa token, fill=oaMuted] at (2.61,1.68) {};
    \node[oa token, fill=oaBlue!48!white] at (3.00,1.68) {};
    \node[oa token, fill=oaMuted] at (3.39,1.68) {};
    \node[oa token, fill=oaBlue!34!white] at (3.78,1.68) {};
    \node[oa token, fill=oaMuted] at (4.17,1.68) {};
    \node[oa token, fill=oaMuted] at (4.56,1.68) {};

    \foreach \x in {2.61,3.39,4.17,4.56}{
      \draw[draw=oaSlate!55, line width=0.35pt]
        (\x-0.10,1.58) -- (\x+0.10,1.78)
        (\x-0.10,1.78) -- (\x+0.10,1.58);
    }

    \draw[oa arrow] (0.96,1.02) -- (4.64,1.02);
    \node[oa note, anchor=north west] at (0.96,0.94) {early};
    \node[oa note, anchor=north east] at (4.64,0.94) {late};

    \node[oa note, anchor=north west, text width=4.95cm] at (0.28,0.60)
      {Independent masks can leave gaps anywhere.};
  \end{scope}

  \begin{scope}[shift={({2*(\panelw+\panelgap)},0)}]
    \draw[oa panel] (0,0) rectangle (\panelw,\panelh);

    \node[oa title, anchor=west] at (0.28,3.63) {PRIOR};
    \node[oa subtitle, anchor=west] at (0.28,3.02)
      {Each finer level is predicted from the coarser};
    \node[oa subtitle, anchor=west] at (0.28,2.70)
      {one; the residuals are the unexplained encoder};
    \node[oa subtitle, anchor=west] at (0.28,2.38)
      {features from lower-level prediction.};
    \node[oa token, fill=oaBlue!88!white] at (1.94,2.00) {};

    \foreach \x in {1.74,2.14}{
      \node[oa token, fill=oaBluePale] at (\x,1.64) {};
      \fill[oaBlue!78!white, rounded corners=0.35mm]
        (\x-0.18,1.65) rectangle (\x+0.18,1.74);
    }

    \foreach \x in {1.36,1.76,2.16,2.56}{
      \node[oa token, fill=oaBluePale] at (\x,1.28) {};
      \fill[oaBlueSoft!90!white, rounded corners=0.35mm]
        (\x-0.18,1.29) rectangle (\x+0.18,1.38);
    }

    \foreach \x in {0.58,0.96,1.34,1.72,2.10,2.48,2.86,3.24}{
      \node[oa tiny token, fill=oaBluePale] at (\x,0.90) {};
      \fill[oaBlueSoft!85!white, rounded corners=0.30mm]
        (\x-0.14,0.91) rectangle (\x+0.14,0.98);
    }

    \draw[oa arrow] (1.94,1.84) -- (1.94,1.74);
    \draw[oa arrow] (1.94,1.48) -- (1.94,1.38);
    \draw[oa arrow] (1.94,1.12) -- (1.94,1.02);

    \draw[oa arrow] (3.54,1.79) -- (1.98,1.79);
    \node[oa callout, anchor=west] at (3.62,1.79)
      {prediction};

    \draw[oa warm arrow] (3.48,1.20) -- (2.84,1.32);
    \node[oa callout, anchor=west, text=oaWarm!90!black] at (3.84,1.18)
      {stored residual};

    \node[oa note, anchor=north west, text width=4.95cm] at (0.28,0.60)
      {Budgets add whole levels: $1,3,7,15,\dots$};
  \end{scope}
\end{tikzpicture}}
\caption{Visual summary of the three bottleneck architectures.}
\label{fig:ordered-architectures}
\end{figure}

For all bottlenecks, we denote the latent representation as $\mathbf{Z}$, an arrangement of $N$ tokens with channel size $C$.
MBOP methods use a flat token sequence, i.e., an $\left[N,C\right]$ tensor;
PRIOR arranges tokens into a pyramidal hierarchy.
Fig.~\ref{fig:ordered-architectures} summarizes the architectural distinction between MBOP-TD, MBOP-ITD, and PRIOR.

\subsection{MBOP}
We consider the forward pass of a single training sample and let $i$ index tokens in $\mathbf{Z}$.
The probability of token $i$ being masked is denoted by $m_i$.
TD samples a single cutoff $c$ from a log-uniform distribution:
\begin{equation}
u \sim \mathcal{U}(0,\log(N+1)), \qquad
c = \lfloor \exp(u) \rfloor, \qquad
m_i = \mathbf{1}[i \ge c].
\end{equation}
ITD instead samples masks independently for each token:
\begin{equation}
m_i \sim \mathrm{Bernoulli}\!\left(\frac{\log(i+1)}{\log(N+1)}\right).
\end{equation}

Besides serving as an additional baseline, ITD allows us to test specific hypotheses within the MBOP framework.
With TD, late tokens face a double disadvantage:
they are updated less often, and their contribution to the loss may remain small because downstream components rely on earlier, more strongly trained tokens.
ITD may improve attribution by creating low-probability events in which late tokens have increased downstream influence.
At the same time, ITD may inadvertently increase redundancy, because conditioning on preceding tokens is weakened.

\subsection{PRIOR}
In PRIOR, $\mathbf{Z}$ is decomposed into $L$ levels with doubling sequence lengths:
\begin{equation}
\sum_{\ell=0}^{L-1} 2^\ell = N .
\end{equation}
Let $\mathbf{y}_\ell$, a tensor of shape $\left[2^\ell, C\right]$, be the unprocessed encoded input to PRIOR at level $\ell$.
From coarser to finer levels, predictor modules $f_{\ell-1}$ refine the latent approximation $\hat{\mathbf{y}}_{\ell-1}$ from level ${\ell-1}$. The residual error $\mathbf{r}_\ell$ is compressed through a token bottleneck $\mathcal{Q}_\ell$. The compressed result is then added to the prediction $\hat{\mathbf{y}}^{\mathrm{pred}}_\ell$, yielding $\hat{\mathbf{y}}_\ell$, which serves as both the level output and the input to the next refinement stage:
\begin{equation}
\hat{\mathbf{y}}^{\mathrm{pred}}_\ell = f_{\ell-1}(\hat{\mathbf{y}}_{\ell-1}), \qquad
\mathbf{r}_\ell = \mathbf{y}_\ell - \hat{\mathbf{y}}^{\mathrm{pred}}_\ell, \qquad
\hat{\mathbf{y}}_\ell = \hat{\mathbf{y}}^{\mathrm{pred}}_\ell + \mathcal{Q}_\ell(\mathbf{r}_\ell).
\end{equation}
For $\ell=0$, the predictor output is zero.
Refinement predictions are trained with
\begin{equation}
\mathcal{R}_{\mathrm{pred}}
=
\frac{1}{L-1}\sum_{\ell=1}^{L-1}
\left\lVert \hat{\mathbf{y}}^{\mathrm{pred}}_\ell - \mathrm{sg}[\mathbf{y}_\ell] \right\rVert_2^2,
\end{equation}
where $\mathrm{sg}$ (stop-gradient) prevents the encoder from adjusting to the predictor.

Each level output $\hat{\mathbf{y}}_\ell$ serves a separate downstream head with objective $\mathcal{J}^{(\ell)}$.
Normalized geometric weighting yields the scalar loss:
\begin{equation}
\mathcal{J}_{\mathrm{PRIOR}}
=
\frac{1}{Z}\sum_{\ell=0}^{L-1} 2^{-(L-1-\ell)} \mathcal{J}^{(\ell)},
\qquad
Z=\sum_{k=0}^{L-1} 2^{-k}.
\end{equation}
This preserves the training signal at all levels, while assigning exponentially larger loss weights to deeper levels, which contain an exponentially larger number of tokens.
In our experiments, all levels use the same objective.

\subsection{Token Parametrizations}
Three variants of tokens are considered in this study: Gaussians, Categoricals, and EMA-VQ codebook vectors.
In the contrastive learning task, tokens are additionally regularized to encourage compression and code usage, as this was found to stabilize performance (cf. Table~\ref{tab:token-parametrizations}).
For PRIOR, parametrized tokens encode the residuals at each level with an added continuous predictor component.
We still consider PRIOR to be a token-bottlenecked model, as the predictor component does not transmit additional sample-specific information beyond the retained residual tokens.

\begin{table}[ht]
\caption{Token parametrizations, masked states, and token-specific regularization terms. $H$ denotes categorical entropy.}
\label{tab:token-parametrizations}
\centering
\small
\setlength{\tabcolsep}{3pt}
\renewcommand{\arraystretch}{1.08}
\begin{tabular*}{\textwidth}{@{\extracolsep{\fill}}lcp{0.62\textwidth}@{}}
\toprule
\mbox{Token Type} & \mbox{Masked State} & Token Regularization \\
\midrule
\mbox{Gaussian} & $\mathcal{N}(\mathbf{0},\mathbf{I})$ & Free-bits KL \cite{kingma2017improvingvariationalinferenceinverse}: $\mathbb{E}_j\left[\max\!\left(0,D_{\mathrm{KL}}-\kappa\right)\right]$ \\
\midrule
\multirow{2}{*}{Categorical} & \multirow{2}{*}{$\frac{1}{C}\cdot\mathbf{1}$} & Sample entropy: $\mathbb{E}\left[H(q)\right]$ \\
& & Negative batch entropy: $-H\left(\mathbb{E}\left[q\right]\right)$ \\
\midrule
\mbox{EMA-VQ} & $\mathbf{0}$ & Commitment: $\mathbb{E}\left[\left\lVert \hat{x}-\mathrm{sg}\!\left[x^{\mathrm{EMA}}\right]\right\rVert_2^2\right]$ \\
\bottomrule
\end{tabular*}
\end{table}

Gaussians are sampled via the reparameterization trick, Categoricals and EMA-VQ use straight-through gradient estimates.
EMA-VQ vectors may select the null state.

\subsubsection{Effective Information}
For each retained set of tokens $S$, we use the validation set to estimate effective information from a token-wise bottleneck usage $b$:
\begin{equation}
I(S)=\sum_{u\in S} b(u)
\end{equation}
Across token types, $b(u)$ estimates the information gained from token $u$ relative to its marginal or inactive state: mutual information for discrete tokens, and the Gaussian variational-rate estimate via KL to the prior \cite{alemi2017deep}:
\begin{equation}
b(u)=
\begin{cases}
\mathbb{E}_{x}\!\left[D_{\mathrm{KL}}\!\left(q_u(z\mid x)\,\|\,p(z)\right)\right] & \quad \text{for Gaussians} \\
\max\!\left(0,H\!\left(\mathbb{E}_{x}\left[q_u(z\mid x)\right]\right)-\mathbb{E}_{x}\left[H\!\left(q_u(z\mid x)\right)\right]\right) & \quad \text{for Categoricals} \\
H\!\left(\mathbb{E}_{x}\left[\operatorname{onehot}(a_u(x))\right]\right) & \quad \text{for EMA-VQ}
\end{cases}
\end{equation}

\subsection{Contrastive Learning}
Training data are sampled from MVImgNet2 \cite{yu2023mvimgnetlargescaledatasetmultiview} object tracks, where separate frames of the same object instance constitute a positive pair.
We use the standard NT-Xent objective \cite{chen2020simple,sohn2016improved}:
\begin{equation}
\ell_i
=
-\log
\frac{
\exp\!\left(s(\mathbf{h}_i,\mathbf{h}_{p(i)})/\tau\right)
}{
\sum_{j=1}^{2B}\mathbf{1}_{j\ne i}\exp\!\left(s(\mathbf{h}_i,\mathbf{h}_j)/\tau\right)
},
\qquad
\mathcal{L}_{\mathrm{NTX}}
=
\frac{1}{2B}\sum_{i=1}^{2B}\ell_i.
\end{equation}
Here, $p(i)$ denotes the positive partner of view $i$, $s$ is cosine similarity, $\tau$ is a temperature parameter, and $B$ is the number of pairs in the batch.
The numerator rewards similarity of encoded positive pairs; the denominator avoids model collapse by penalizing similarities of negative pairs.
As in \cite{chen2020simple}, we use projection heads, i.e., small MLPs producing $\mathbf{h}_i$ as a projected, lower-dimensional descriptor of view $i$.
We adapt this idea to PRIOR by means of level-wise projection heads.

For the encoder, we use FLIP \cite{traub2025lookinglocallyobjectcentricvision}, a transformer-based, object-centric vision model with foveal multi-scale patching to selectively process individual objects.
Regularizer weights were selected by a two-stage Bayesian optimization procedure.
Broad stage-1 samples were first evaluated in short-horizon runs; a Gaussian-process surrogate \cite{rasmussen2006gaussian} then guided stage-2 refinement using strong anchors, posterior-mean candidates, and expected improvement \cite{jones1998efficient}.

MVImgNet2 provides category labels at four taxonomy levels, from abstract to nuanced, referred to here as Coarse-, Mid-, Fine-, and Class-level.
Readouts are evaluated by means of linear probes trained on frozen representations.
Table~\ref{tab:contrastive-parameter-counts} reports the resulting contrastive model sizes, separating shared encoders, projection heads, EMA codebooks, and PRIOR self-predictors.

\begin{table}[ht]
\caption{
Model parameters in millions by architecture, token type and model component.
EMA codebooks are accounted for separately, as they are not directly optimized but track network activity over time and become a component of the representation.
Self-Pred. refers to the parameters of PRIOR's refinement modules.
Total accounts for parameters used in downstream tasks, i.e., without projection heads, with bracketed numbers including them.
}
\label{tab:contrastive-parameter-counts}
\centering
\small
\setlength{\tabcolsep}{3pt}
\renewcommand{\arraystretch}{1.05}
\begin{tabular*}{\textwidth}{@{\extracolsep{\fill}}llrrrrr@{}}
\toprule
Architecture & Token Type & FLIP & Projection & EMA & Self-Pred. & Total \\
\midrule
\multirow{3}{*}{PRIOR} & Gaussian & 3.042 & 0.929 & -- & 3.778 & 6.819 [7.748] \\
& Categorical & 2.926 & 0.929 & -- & 3.778 & 6.704 [7.633] \\
& EMA-VQ & 2.811 & 0.929 & 0.259 & 3.778 & 6.847 [7.776] \\
\midrule
\multirow{3}{*}{MBOP} & Gaussian & 6.996 & 0.133 & -- & -- & 6.996 [7.129] \\
& Categorical & 4.899 & 0.133 & -- & -- & 4.899 [5.032] \\
& EMA-VQ & 2.802 & 0.133 & 4.145 & -- & 6.948 [7.080] \\
\bottomrule
\end{tabular*}

\end{table}

\subsection{Reconstruction}
A wavelet-space vision transformer is trained on FFHQ autoencoding \cite{karras2019stylebasedgeneratorarchitecturegenerative}.
As in contrastive learning, reconstruction with PRIOR uses level-wise decoders.
The parameter count for full-budget reconstruction was approximately $11.479$ million for MBOP models and $11.761$ million for PRIOR.

\newlength{\gaussianreconsize}

\newcommand{\bestmark}{\rlap{$^{\star}$}}
\newcommand{\secondmark}{\rlap{\raisebox{0.90ex}{\scalebox{0.45}{$+$}}}}

\newcommand{\gaussgt}[1]{%
figures/reconstruction_examples/gaussian/gt/img_#1.png%
}

\newcommand{\gaussitdimg}[2]{%
figures/reconstruction_examples/gaussian/itd/#1_img_#2.png%
}

\newcommand{\gausstdimg}[2]{%
figures/reconstruction_examples/gaussian/td/#1_img_#2.png%
}

\newcommand{\gausspriorimg}[2]{%
figures/reconstruction_examples/gaussian/prior/#1_img_#2.png%
}

\newcommand{\gaussianreconimage}[1]{%
\includegraphics[width=\gaussianreconsize]{#1}%
}

\newcommand{\gaussianreconfigure}[1]{%
\begingroup
\setlength{\gaussianreconsize}{0.107\linewidth}%
\begin{tikzpicture}[x=\gaussianreconsize, y=\gaussianreconsize, inner sep=0pt, outer sep=0pt]
\node[anchor=east] at (-0.09,-0.5) {ITD};
\node[anchor=east] at (-0.09,-1.5) {TD};
\node[anchor=east] at (-0.09,-2.5) {PRIOR};
\node[anchor=north west] at (0,0) {\gaussianreconimage{\gaussgt{#1}}};
\node[anchor=north west] at (1,0) {\gaussianreconimage{\gaussitdimg{level_000_tokens_001}{#1}}};
\node[anchor=north west] at (2,0) {\gaussianreconimage{\gaussitdimg{level_002_tokens_003}{#1}}};
\node[anchor=north west] at (3,0) {\gaussianreconimage{\gaussitdimg{level_006_tokens_007}{#1}}};
\node[anchor=north west] at (4,0) {\gaussianreconimage{\gaussitdimg{level_014_tokens_015}{#1}}};
\node[anchor=north west] at (5,0) {\gaussianreconimage{\gaussitdimg{level_030_tokens_031}{#1}}};
\node[anchor=north west] at (6,0) {\gaussianreconimage{\gaussitdimg{level_062_tokens_063}{#1}}};
\node[anchor=north west] at (7,0) {\gaussianreconimage{\gaussitdimg{level_126_tokens_127}{#1}}};
\node[anchor=north west] at (0,-1) {\gaussianreconimage{\gaussgt{#1}}};
\node[anchor=north west] at (1,-1) {\gaussianreconimage{\gausstdimg{level_000_tokens_001}{#1}}};
\node[anchor=north west] at (2,-1) {\gaussianreconimage{\gausstdimg{level_002_tokens_003}{#1}}};
\node[anchor=north west] at (3,-1) {\gaussianreconimage{\gausstdimg{level_006_tokens_007}{#1}}};
\node[anchor=north west] at (4,-1) {\gaussianreconimage{\gausstdimg{level_014_tokens_015}{#1}}};
\node[anchor=north west] at (5,-1) {\gaussianreconimage{\gausstdimg{level_030_tokens_031}{#1}}};
\node[anchor=north west] at (6,-1) {\gaussianreconimage{\gausstdimg{level_062_tokens_063}{#1}}};
\node[anchor=north west] at (7,-1) {\gaussianreconimage{\gausstdimg{level_126_tokens_127}{#1}}};
\node[anchor=north west] at (0,-2) {\gaussianreconimage{\gaussgt{#1}}};
\node[anchor=north west] at (1,-2) {\gaussianreconimage{\gausspriorimg{level_000_tokens_001}{#1}}};
\node[anchor=north west] at (2,-2) {\gaussianreconimage{\gausspriorimg{level_001_tokens_003}{#1}}};
\node[anchor=north west] at (3,-2) {\gaussianreconimage{\gausspriorimg{level_002_tokens_007}{#1}}};
\node[anchor=north west] at (4,-2) {\gaussianreconimage{\gausspriorimg{level_003_tokens_015}{#1}}};
\node[anchor=north west] at (5,-2) {\gaussianreconimage{\gausspriorimg{level_004_tokens_031}{#1}}};
\node[anchor=north west] at (6,-2) {\gaussianreconimage{\gausspriorimg{level_005_tokens_063}{#1}}};
\node[anchor=north west] at (7,-2) {\gaussianreconimage{\gausspriorimg{level_006_tokens_127}{#1}}};
\node[anchor=north, yshift=-1.0ex] at (0.5,-3) {GT};
\node[anchor=north, yshift=-1.0ex] at (1.5,-3) {1};
\node[anchor=north, yshift=-1.0ex] at (2.5,-3) {3};
\node[anchor=north, yshift=-1.0ex] at (3.5,-3) {7};
\node[anchor=north, yshift=-1.0ex] at (4.5,-3) {15};
\node[anchor=north, yshift=-1.0ex] at (5.5,-3) {31};
\node[anchor=north, yshift=-1.0ex] at (6.5,-3) {63};
\node[anchor=north, yshift=-1.0ex] at (7.5,-3) {127};
\end{tikzpicture}%
\endgroup
}

\section{Results}
We evaluate ordered bottlenecks according to two criteria:
whether they preserve strong peak performance, and whether additional tokens provide complementary, fine-grained information.

\subsection{Contrastive Learning of Ordered Object Codes}
Before analyzing frozen linear probes, Table~\ref{tab:contrastive-simclr-loss} reports the validation NT-Xent losses of the trained contrastive models.
PRIOR reaches the lowest validation loss in all token families; among MBOP models, ITD is lower for Categorical and EMA-VQ, and TD for Gaussian.

\begin{table}[ht]
\centering
\caption{NT-Xent validation loss of the trained models. Bold marks the lowest loss within the token family. PRIOR models report the loss value from the head of the final level.}
\label{tab:contrastive-simclr-loss}
\small
\setlength{\tabcolsep}{8pt}
\begin{tabular}{@{}lccc@{}}
	\toprule
	Token type & PRIOR opt. & MBOP-ITD opt. (match) & MBOP-TD opt. (match) \\
	\midrule
	Gaussian & \textbf{1.244} & 1.337 (1.391) & 1.288 (1.424) \\
	Categorical & \textbf{1.400} & 1.844 (2.137) & 2.881 (2.664) \\
	EMA-VQ & \textbf{1.350} & 1.426 (1.994) & 2.183 (2.376) \\
\bottomrule
\end{tabular}
\end{table}

\begin{figure}[ht]
\centering
\begin{minipage}[t]{0.46\linewidth}
\centering
\includegraphics[width=0.90\linewidth]{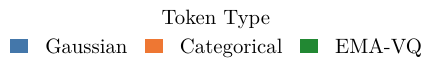}\\[-0.4ex]
\includegraphics[width=\linewidth]{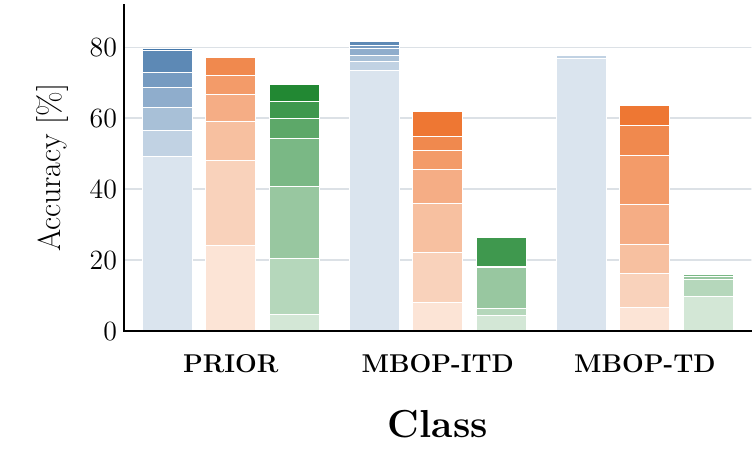}
\end{minipage}
\hfill
\begin{minipage}[t]{0.46\linewidth}
\centering
\includegraphics[width=0.96\linewidth]{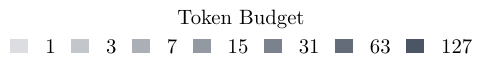}\\[-0.4ex]
\includegraphics[width=\linewidth]{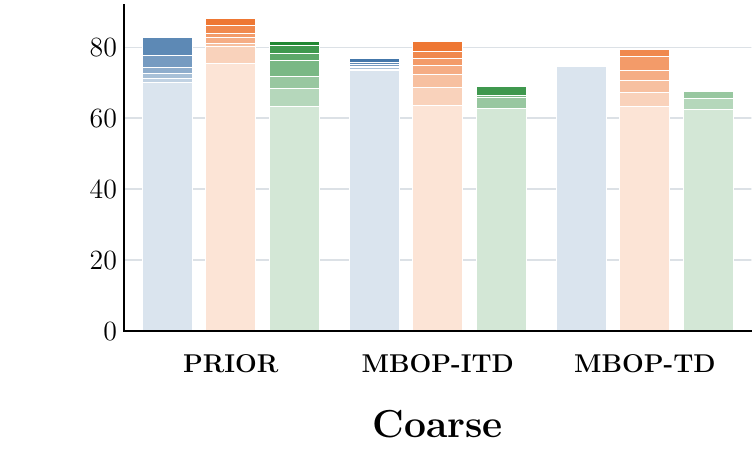}
\end{minipage}
\caption{
	Budget-wise linear-probe accuracy for class and coarse taxonomy levels.
	Shading decomposes readout quality by utilized tokens up to the highest-performing prefix of the respective model;
	for PRIOR, a prefix corresponds to the first $k$ concatenated levels.
}
\label{fig:contrastive-budget-allocation}
\end{figure}

\begin{table}[ht]
\centering
\begingroup
\scriptsize
\setlength{\tabcolsep}{1.8pt}
\renewcommand{\arraystretch}{0.88}
\caption{
Frozen linear-probe accuracy across architectures, tokens, budgets, and taxonomy levels.
Budgets describe (level-) prefixes; bold and underlining indicate the best and second-best model within each family, budget, and taxonomy level.
Star and plus mark the best and second-best model within each family and taxonomy level, \textit{across} budgets.
}
\label{tab:taxonomy-budget-accuracy}
\resizebox{\textwidth}{!}{%
\begin{tabular}{@{}l r@{\hspace{0.60em}} r@{\hspace{0.60em}} r@{\hspace{0.60em}} r@{\hspace{0.60em}} r@{\hspace{0.60em}} r@{\hspace{0.60em}} r@{\hspace{0.60em}} r@{\hspace{0.60em}} r@{\hspace{0.60em}} r@{\hspace{0.60em}} r@{\hspace{0.60em}} r@{\hspace{0.60em}}@{}}
\toprule
& \multicolumn{4}{c}{PRIOR} & \multicolumn{4}{c}{MBOP-ITD} & \multicolumn{4}{c}{MBOP-TD} \\
\cmidrule(lr){2-5}\cmidrule(lr){6-9}\cmidrule(l){10-13}
Budget & Coarse & Mid & Fine & Class & Coarse & Mid & Fine & Class & Coarse & Mid & Fine & Class \\
\midrule
\multicolumn{13}{@{}l}{\textbf{Gaussian}} \\
1 & 70.11 & 60.46 & 44.60 & 49.23 & \underline{73.35} & \underline{63.95} & \underline{51.68} & \underline{73.49} & \textbf{74.45} & \textbf{65.23} & \textbf{53.10} & \textbf{76.91} \\
3 & 71.22 & 61.46 & 46.05 & 56.45 & \underline{73.75} & \underline{64.67} & \underline{52.27} & \underline{76.12} & \textbf{74.62} & \textbf{66.25} & \textbf{53.97} & \textbf{77.60} \\
7 & 72.46 & 63.19 & 48.67 & 62.91 & \underline{74.20} & \underline{65.56} & \underline{53.32} & \textbf{77.72} & \textbf{74.94} & \textbf{65.95} & \textbf{54.16} & \underline{77.64} \\
15 & 74.32 & 65.33 & 51.88 & 68.71 & \underline{74.60} & \textbf{66.09} & \underline{54.05} & \textbf{79.70} & \textbf{74.71} & \underline{66.00} & \textbf{54.18} & \underline{77.75} \\
31 & \textbf{77.55} & \textbf{68.82} & \textbf{56.17} & 73.01 & \underline{75.07} & \underline{66.64} & \underline{55.26} & \textbf{80.58} & 74.76 & 66.06 & 54.28 & \underline{77.75} \\
63 & \textbf{82.70} & \textbf{73.52} & \textbf{63.65} & \underline{78.95} & \underline{75.75} & \underline{67.57} & \underline{56.72} & \textbf{81.75}\bestmark & 74.80 & 65.53 & 54.12 & 77.95 \\
127 & \textbf{82.94}\bestmark & \textbf{75.46}\bestmark & \textbf{64.97}\bestmark & \underline{79.74}\secondmark & \underline{76.82}\secondmark & \underline{68.78}\secondmark & \underline{58.57}\secondmark & \textbf{80.30} & 74.80 & 66.23 & 54.53 & 77.35 \\
\midrule
\multicolumn{13}{@{}l}{\textbf{Categorical}} \\
1 & \textbf{75.37} & \textbf{65.42} & \textbf{48.32} & \textbf{24.16} & \underline{63.66} & \underline{54.03} & \underline{36.49} & \underline{7.96} & 63.22 & 53.68 & 36.38 & 6.48 \\
3 & \textbf{80.27} & \textbf{71.05} & \textbf{56.14} & \textbf{48.17} & \underline{68.70} & \underline{59.02} & \underline{42.25} & \underline{22.11} & 67.36 & 57.67 & 40.99 & 16.28 \\
7 & \textbf{81.14} & \textbf{72.44} & \textbf{58.45} & \textbf{59.11} & \underline{72.35} & \underline{62.85} & \underline{47.10} & \underline{35.91} & 70.66 & 61.16 & 44.42 & 24.34 \\
15 & \textbf{82.87} & \textbf{74.41} & \textbf{61.67} & \textbf{66.63} & \underline{74.92} & \underline{65.58} & \underline{51.08} & \underline{45.41} & 73.49 & 64.05 & 48.01 & 35.61 \\
31 & \textbf{83.85} & \textbf{75.95} & \textbf{64.88} & \textbf{71.94} & 76.83 & 67.56 & 53.64 & \underline{50.92} & \underline{77.42} & \underline{68.16} & \underline{54.33} & 49.50 \\
63 & \textbf{86.07} & \textbf{79.51} & \textbf{68.43} & \textbf{77.17}\bestmark & 78.89 & 70.45 & 56.59 & 54.91 & \underline{79.45} & \underline{70.81} & \underline{57.88} & \underline{58.02} \\
127 & \textbf{88.01}\bestmark & \textbf{80.45}\bestmark & \textbf{70.37}\bestmark & \textbf{75.81} & \underline{81.73}\secondmark & \underline{73.40}\secondmark & \underline{60.97}\secondmark & 61.87 & 79.73 & 71.32 & 58.51 & \underline{63.67}\secondmark \\
\midrule
\multicolumn{13}{@{}l}{\textbf{EMA-VQ}} \\
1 & \textbf{63.28} & \textbf{52.93} & \textbf{35.03} & \underline{4.55} & \underline{62.83} & \underline{52.68} & \underline{34.74} & 4.37 & 62.50 & 51.68 & 27.87 & \textbf{9.74} \\
3 & \textbf{68.44} & \textbf{58.68} & \textbf{40.78} & \textbf{20.50} & 62.79 & \underline{52.51} & \underline{34.75} & 6.18 & \underline{65.50} & 52.25 & 33.28 & \underline{14.40} \\
7 & \textbf{71.85} & \textbf{61.80} & \textbf{45.82} & \textbf{40.71} & 65.87 & \underline{56.17} & \underline{39.08} & \underline{17.75} & \underline{67.42} & 56.15 & 31.61 & 15.35 \\
15 & \textbf{76.21} & \textbf{67.16} & \textbf{52.68} & \textbf{54.16} & 66.52 & \underline{56.45} & \underline{39.22} & \underline{18.06} & \underline{67.59} & 55.82 & 34.82 & 15.90 \\
31 & \textbf{78.16} & \textbf{69.83} & \textbf{56.01} & \textbf{60.03} & 66.48 & 56.86 & \underline{39.43} & \underline{17.30} & \underline{67.01} & \underline{57.00} & 35.54 & 15.00 \\
63 & \textbf{80.50} & \textbf{70.98} & \textbf{60.01} & \textbf{64.63} & \underline{69.02}\secondmark & \underline{58.56}\secondmark & \underline{40.73}\secondmark & \underline{26.37}\secondmark & 67.21 & 54.57 & 35.63 & 14.91 \\
127 & \textbf{81.57}\bestmark & \textbf{73.31}\bestmark & \textbf{61.86}\bestmark & \textbf{69.42}\bestmark & \underline{67.33} & 56.72 & \underline{37.05} & \underline{24.97} & 66.75 & \underline{57.25} & 35.63 & 13.77 \\
\bottomrule
\end{tabular}%
}
\endgroup
\end{table}

\begin{figure}[ht]
\centering
\includegraphics[width=\linewidth]{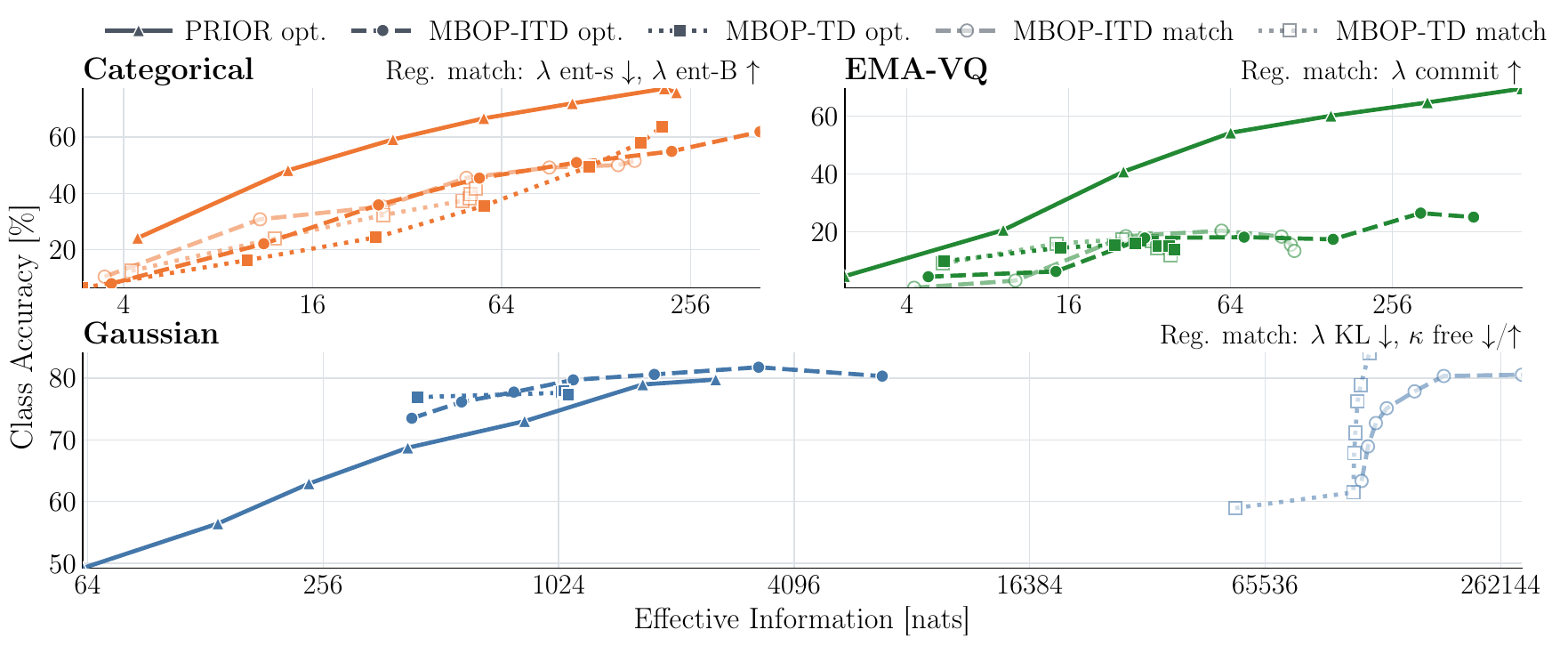}
\caption{
	Class accuracy versus log-scaled effective information for retained prefix budgets across tuned architectures and regularizer-matched controls.
	Arrows indicate directional changes of the respective weights in the control setting.
	Single arrows indicate both variants, two arrows indicate ITD/TD changes.
}
\label{fig:contrastive-effective-information}
\end{figure}

Fig.~\ref{fig:contrastive-budget-allocation} summarizes budget-wise linear-probe accuracy for class and coarse taxonomy readouts;
complete results are given in Table~\ref{tab:taxonomy-budget-accuracy}.
Differences between MBOP-ITD and MBOP-TD tend to be small, with a tendency for higher accuracies in MBOP-ITD.
EMA-VQ models are the exception; ITD strongly improves over TD (e.g., peak class accuracy of $26.37\%$ vs. $15.9\%$).
As these models were tuned individually to generate maximally strong baselines against PRIOR with respect to their training objective, ITD--TD results on their own do not grant clear attribution to masking schemes.
Therefore, we conducted a control experiment in which each ITD--TD pair was tuned to optimize the combined, average performance.
Within the control, MBOP differences were similar with respect to directionality and magnitude (cf. Fig.~\ref{fig:contrastive-effective-information}).

Conversely, PRIOR and MBOP baselines differ strongly, both with respect to peak accuracies and in relation to ordering.
Highest accuracies per model and token parametrizations are predominantly observed with PRIOR.
The exception is Gaussian class accuracy, where MBOP-ITD peaks at $81.75\%$, followed by PRIOR at $79.74\%$ and MBOP-TD at $77.95\%$.
Coarse, mid, and fine readouts are all maximized by PRIOR, followed by MBOP-ITD.
For instance, within the Gaussian family at the mid level, PRIOR reaches $75.46\%$, MBOP-ITD $68.78\%$, and MBOP-TD $66.25\%$.

Particularly strong differences are observed in the Categorical and EMA-VQ families.
For example, Categorical-PRIOR reaches a class accuracy of $77.17\%$ versus $63.67\%$ for MBOP-TD, and EMA-VQ-PRIOR peaks at $69.42\%$ compared to MBOP-ITD at only $26.37\%$.

Considering peak accuracies, PRIOR generally improves over MBOP within token families.
More importantly, peak accuracies fall into a tighter range, lifting discrete and quantized tokens much closer to the Gaussian ceiling.
As representational capacity across models differs significantly, Fig.~\ref{fig:contrastive-effective-information} controls for effective information $I$.
Accuracy differences in the discrete and quantized settings remain high, making mere bandwidth differences unlikely as a viable explanation.

PRIOR shows robust ordering of fine-grained semantic information as measured by readouts.
Across token families, class-level accuracies strongly improve with increased budgets.
Gaussian PRIOR, for instance, improves from $49.23\%$ at one token to $79.74\%$ at full budget;
EMA-VQ PRIOR does so from $4.55\%$ to $69.42\%$.
Qualitatively, budget-dependent differences are more evenly distributed and stepwise in PRIOR models (cf. Fig.~\ref{fig:contrastive-budget-allocation}, Fig.~\ref{fig:contrastive-effective-information}).

Adjusting for their weaker ceiling performance, discrete and quantized MBOP models show similarly strong class-level ordering effects.
However, their Gaussian ITD counterpart is only weakly ordered ($73.49\%-81.75\%$), and TD is virtually flat ($76.91\%-77.95\%$).
The control experiments show that lower regularization can improve ordering in Gaussian MBOP.
However, the contrastive training objective is suboptimal in these cases and, more importantly, effective information is exponentially higher (cf. Fig.~\ref{fig:contrastive-effective-information}, Table~\ref{tab:contrastive-simclr-loss}).

Because of their strong but quickly saturating performance, Gaussian MBOP models tend to outperform Gaussian PRIOR within the first budget quantile (up to $31$ tokens), after which PRIOR converts additional budget into stronger readouts. Within the Categorical and EMA-VQ families, PRIOR yields stronger readouts across nearly all budgets;
the one exception is EMA-VQ class readout with a single token.

The contrastive experiments can therefore be summarized as follows:
Compared against baselines, PRIOR yields strong, coarse-to-fine ordered representations without exception.
It most clearly improves discrete and quantized representations where MBOP models are comparatively weak, especially with EMA-VQ.
With Gaussian tokens, accuracy differences are small and inconsistent;
PRIOR improves mainly the budget profiles.

\subsection{Ordered Variational Autoencoding}
\begin{figure}[ht]
\centering
\includegraphics[width=\linewidth]{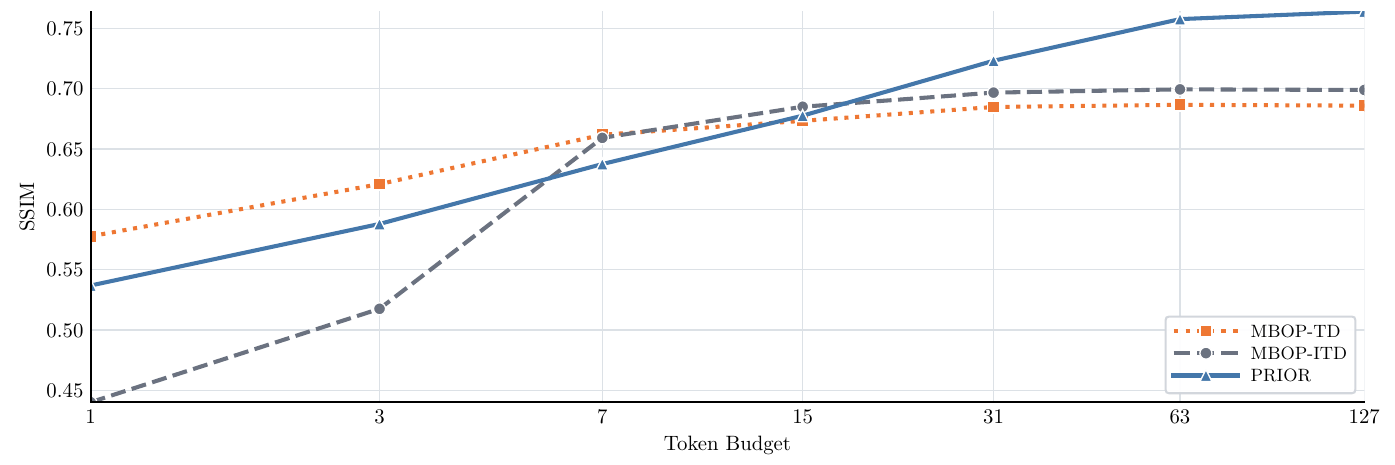}
\caption{Mean validation SSIM ($\uparrow$) across cumulative token budgets.}
\label{fig:ffhq-recon-ssim-gaussian}
\end{figure}

Fig.~\ref{fig:ffhq-recon-ssim-gaussian} and Table~\ref{tab:ffhq-recon-metrics-gaussian} show a complementary pattern for FFHQ reconstruction.
MBOP-TD is strongest at the smallest budgets, while MBOP-ITD is weakest.
From $15$ tokens onward, MBOP-ITD overtakes MBOP-TD, but both variants saturate early.
In contrast, PRIOR improves steadily as additional levels become active, surpasses both MBOP variants from $31$ tokens onward and continues improving up to the full budget.
Representative examples in Fig.~\ref{fig:ffhq-recon-qualitative-main} are consistent with this result:
TD provides plausible low-budget reconstruction but changes little later; ITD fails at the one-token extreme; and PRIOR adds visible detail across successive levels.

\begin{figure}[ht]
\centering
\gaussianreconfigure{00}
\caption{Representative ground-truth (GT) images and reconstructions across architectures and token budgets.}
\label{fig:ffhq-recon-qualitative-main}
\end{figure}

\begin{table}[t]
\caption{FFHQ reconstruction validation means across token budgets. Best values are bold; second-best values are underlined.}
\centering
\scriptsize
\setlength{\tabcolsep}{3.2pt}
\renewcommand{\arraystretch}{1.06}
\begin{tabular*}{\textwidth}{@{\extracolsep{\fill}}llcccc@{}}
\toprule
Metric & Setup & 1 & 15 & 31 & 127 \\
\midrule
SSIM ($\uparrow$) & PRIOR & \underline{0.5370} & \underline{0.6777} & \textbf{0.7232} & \textbf{0.7639} \\
 & MBOP-ITD & 0.4403 & \textbf{0.6851} & \underline{0.6968} & \underline{0.6990} \\
 & MBOP-TD & \textbf{0.5778} & 0.6734 & 0.6849 & 0.6860 \\
\addlinespace[0.75ex]
L1 ($\downarrow$) & PRIOR & \underline{0.0833} & \underline{0.0420} & \textbf{0.0348} & \textbf{0.0301} \\
 & MBOP-ITD & 0.1716 & \textbf{0.0411} & \underline{0.0385} & \underline{0.0379} \\
 & MBOP-TD & \textbf{0.0725} & 0.0440 & 0.0419 & 0.0419 \\
\addlinespace[0.75ex]
L2 ($\downarrow$) & PRIOR & \underline{0.0143} & \underline{0.0044} & \textbf{0.0031} & \textbf{0.0023} \\
 & MBOP-ITD & 0.0585 & \textbf{0.0043} & \underline{0.0038} & \underline{0.0037} \\
 & MBOP-TD & \textbf{0.0116} & 0.0047 & 0.0043 & 0.0043 \\
\addlinespace[0.75ex]
MS-SSIM ($\uparrow$) & PRIOR & \underline{0.7156} & \underline{0.9184} & \textbf{0.9505} & \textbf{0.9699} \\
 & MBOP-ITD & 0.4882 & \textbf{0.9227} & \underline{0.9348} & \underline{0.9366} \\
 & MBOP-TD & \textbf{0.7669} & 0.9127 & 0.9247 & 0.9250 \\
\bottomrule
\end{tabular*}
\label{tab:ffhq-recon-metrics-gaussian}
\end{table}

\section{Discussion}
The results suggest that PRIOR is a promising framework for ordered representation learning.
Across several experimental settings, we compared PRIOR against strong baselines.
In these experiments, PRIOR was the only model that consistently approached the main goal of ordered representations: learning coarse-to-fine codes while maintaining strong peak performance.

For MBOP, our results indicate a tension between prefix performance and overall representation quality.
At high capacity, MBOP can use early tokens very efficiently, but it does not produce a clearly ordered semantic allocation.
Settings that improve ordering tend to reduce peak performance.
The results suggest that the forms of ordering induced by MBOP constrain capacity or lead to underfitting, rather than inducing robust structural organization.

Independent masking does not fully resolve this tension in our experiments.
MBOP-ITD sometimes improves over MBOP-TD, especially at moderate to high budgets, but the gains are comparatively small and less consistent than the differences between PRIOR and MBOP.

Overall, the results support the main hypotheses that motivated PRIOR.
Strong coarse readouts from early levels, together with consistent late-level improvements in detailed readouts, suggest that self-predictive residual refinement provides an effective ordering mechanism.
The pronounced utility of late tokens in discrete and quantized models further supports the view that restricted gradient exposure is a limiting factor for MBOP, but not for PRIOR.
The analysis of effective information is consistent with this interpretation, indicating that PRIOR-induced orderings can transmit information more efficiently.

Taken together, these findings point to several directions for future work.
Further evaluations should assess the breadth of PRIOR's applicability, the robustness of learned hierarchical encodings, and the semantic nature of these encodings.
PRIOR also offers additional design choices that were not exhaustively studied here, such as alternative weighting schemes or different per-level objectives.
Most importantly, we intend to embed PRIOR into larger architectures and broader problem settings.
PRIOR's emergent hierarchical predictive structure enables refinement ``on demand''.
We believe that this ability could be especially useful for developing hierarchical world models that can be flexibly recruited for task-adaptive perception and planning.
Such models may be particularly powerful in environments where relevant objects and interactions change depending on the task and environmental context.

\begin{credits}
\subsubsection{\ackname}
We acknowledge funding by the Deutsche Forschungsgemeinschaft (DFG, German Research Foundation) -- 381713393, 467045002, 564829065.
Martin Butz is a member of the Machine Learning Cluster of Excellence -- 390727645.
The authors thank the International Max Planck Research School for Intelligent Systems (IMPRS-IS) for supporting Manuel Traub.

\subsubsection{\discintname}
The authors have no competing interests to declare that are relevant to the content of this article.
\end{credits}

\bibliographystyle{splncs04}
\bibliography{literature}

\end{document}